\documentclass{article} 
\usepackage{iclr2027_conference,times}

\usepackage{caption}
\usepackage{wrapfig}
\usepackage{hyperref}
\usepackage{url}
\usepackage{amsmath,amssymb,amsthm}
\usepackage{graphicx}        
\usepackage{booktabs}        
\usepackage{multirow}        
\usepackage{xcolor}          
\usepackage{colortbl}
\usepackage{microtype}       
\usepackage{xspace}          
\usepackage{cleveref}        
\crefname{figure}{Fig.}{Figs.}
\Crefname{figure}{Fig.}{Figs.}
\crefname{table}{Tab.}{Tabs.}
\Crefname{table}{Tab.}{Tabs.}
\crefname{section}{Sec.}{Secs.}
\Crefname{section}{Sec.}{Secs.}
\crefname{subsection}{Sec.}{Secs.}
\Crefname{subsection}{Sec.}{Secs.}
\crefname{equation}{Eq.}{Eqs.}
\Crefname{equation}{Eq.}{Eqs.}
\usepackage{pifont}          
\usepackage{enumitem}        
\usepackage[normalem]{ulem}  

\newcommand{\circnum}[1]{\raisebox{-1.1pt}{\ding{\numexpr181+#1\relax}}}
\newcommand{\imp}[2]{\begin{tabular}[t]{@{}c@{}}\textbf{#1}\\[-2pt]{\scriptsize +#2\%}\end{tabular}}

\definecolor{citecolor}{RGB}{30, 90, 170}   
\definecolor{linkcolor}{RGB}{170, 40, 40}   
\hypersetup{
  colorlinks = true,
  citecolor  = citecolor,
  linkcolor  = linkcolor,
  urlcolor   = citecolor,
}

\setcitestyle{numbers,square,comma}

\newcommand{\method}{\textbf{\textsc{SafeHarness}}\xspace}     

\newcommand{\bench}{SafeLIBERO\xspace}
\newcommand{\harness}{Harness VLA\xspace}

\newcommand{\pifive}{$\pi_{0.5}$\xspace}

\newcommand{\tsr}{TSR\xspace}
\newcommand{\car}{CAR\xspace}
\newcommand{\ssr}{SSR\xspace}

\theoremstyle{plain}

\title{SafeHarness: Coding Agents with Harness for Safe Robot Control}

\author{Bingxin Xu$^{1}$ \quad Yuzhang Shang$^{2}$ \quad Zhen Dong$^{3}$ \quad Emilio Ferrara$^{1}$ \\
$^{1}$USC \quad $^{2}$UCF \quad $^{3}$UCSB
}

 \iclrfinalcopy 

\begin{document}

\maketitle

\begin{abstract}
Coding agents have emerged as a promising paradigm for robot
manipulation: a language model writes the robot controller as a
program, and agents built in this way now operate robots without
robot-specific training.
Whether this paradigm is also safe, however, has not been asked.
We evaluate coding agent under a safety constraint, where
each task pairs a manipulation goal with an obstacle the robot must
not touch.
The agent pursues the goal but collides with the obstacle in most
cases, treating task completion as its sole objective while
neglecting safety.
The agent reasons about the obstacle in its traces, and the prompt already forbids touching it, so neither perception nor instruction is at fault; the fault lies in the planning, where the stated constraint never becomes a priority.
By decomposing manipulation into a route phase and a contact-rich moment, we locate the source of the failure.
Along the route, the model cannot prioritize the safety constraint,
having no notion of a clearing route and none of replanning once a
chosen route becomes infeasible.
At the contact, it is unaware that contact execution is bounded by the
same constraint.
To close this gap, we present \method, which equips the model with two
obstacle-aware harnesses that enable it to prioritize the safety
constraint.
Obstacle-aware route planning grounds the objects as bounding boxes
and draws candidate routes over them as sequences of waypoints.
The agent then plans a route in advance, verifies it, replans when
necessary, and only then executes it.
Obstacle-aware contact execution instead selects the contact position
so that the contact itself avoids the obstacle.
\method attains $81.2\%$ task success and $91.9\%$ collision avoidance with GPT-6-Astra, surpassing the previous SOTA by $13.7$ and $23.0$ points, respectively. Compared with the same agent without harnesses, these results represent gains of $31.2$ and $57.5$ points, respectively.
\end{abstract}

\section{Introduction}
\label{sec:intro}

Coding agents have emerged as a promising paradigm for robot manipulation. Large Language Models (LLMs) can write robot controllers as programs.
Code as Policies \citep{liang2023code} set the pattern: the model
composes perception and control APIs into a program, and the program
is the policy.
Later agents rewrite their own controller code after a failure
\citep{kumar2026actobserve}, keep a library of skills that worked
\citep{meng2025growing}, and spend more test-time compute for more
reliability \citep{fu2026capx}.
Frontier agents now drive real robots without robot-specific training
\citep{jia2026agentpolicy,xiao2026enpire}.
Since the release of GPT-6 Astra \citep{openai2026astra}, many groups
have reported promising robot results with it.
With no robot-specific training, Astra completes pick-and-place on a
real arm almost perfectly \citep{robocurve2026astra,jia2026agentpolicy},
and paired with a frozen \pifive{} it extends to bimanual,
contact-rich tasks \citep{su2026astra}.
Yet Astra contains no robot policy.
Every robot result attributed to it belongs to the harness around it,
the tools the agent is given to call, and coding agents make this
division of labor explicit.
\harness \citep{zhang2026harness} lets the agent explore a scene, keep
what worked as a skill memory, and reuse it on later tasks, calling a
frozen VLA only where the task turns contact-rich.
\citet{xu2026baton} extends it to long horizons.

None of this work has asked whether such paradigm on robot manipulation is safe.
The studies above score a rollout on whether the goal was reached, and
the benchmarks they inherit do the same
\citep{liu2023libero,cui2026liberosafety}.
What the robot touched on the way is never measured.
Yet a collision in a cluttered workspace damages hardware, injures
people, and destroys property \citep{haddadin2017collisions}.
Success does not imply safety \citep{huang2026safemanip}.
Across a model generation the two can even diverge
\citep{chen2026hazardarena,fan2026safevlabench}.
Scaling up data does not resolve this either.
Training on ten times as many collision-free
demonstrations lowers the collision rate by under three points
\citep{cui2026liberosafety}.
Completing the task and touching nothing else are two halves of one
requirement.

\begin{figure}[t]
  \centering
  \includegraphics[width=0.99\textwidth]{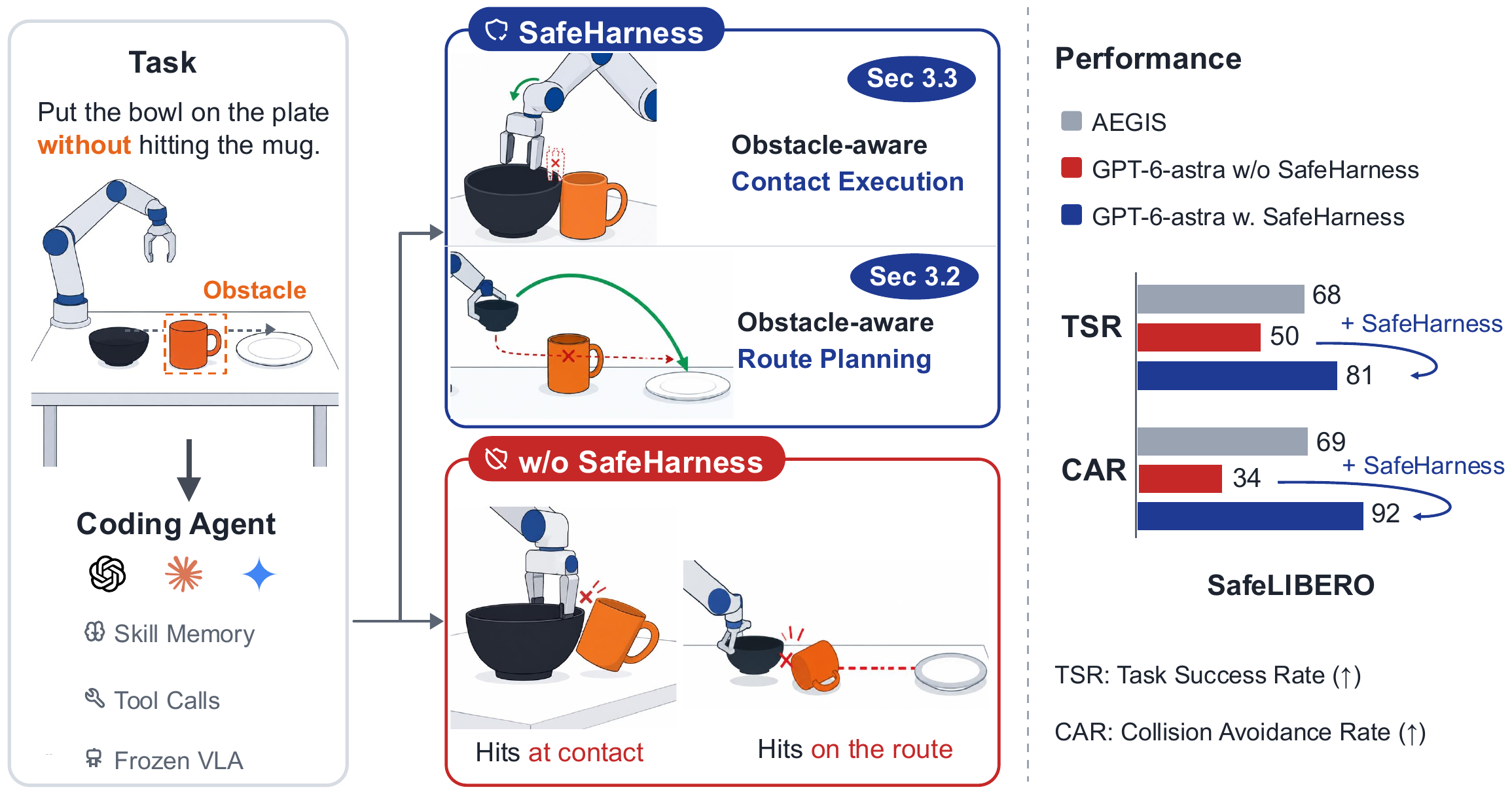}
  \caption{\textbf{\method puts safety on par with task completion.} 
  Coding agent receives a task instruction with a safety constraint, along with observation images. \method then equips it with obstacle-aware route planning (\cref{sec:route_planning}) and obstacle-aware contact execution (\cref{sec:contact_execution}) to enforce safety throughout the task.}
  \vspace{-0.3in}
  \label{fig:teaser}
\end{figure}

To solve this problem, we first evaluate the coding agent on robot safety benchmark. Upon evaluation, \underline{\textit{we observe that coding agent performs poorly under a safety constraint}}.
The coding agent completes
the task but collides with the obstacle in most cases.
It treats the task goal as the core objective and neglects that safety
has to be satisfied as well.
The cause is not recognition, since the agent identifies the obstacle
correctly.
Nor is it instruction.
Adding the safety requirement to the prompt does not change the
behavior \citep{lu2026isbench,yang2026saferelbench}: the agent restates
the constraint and then violates it.
The reasons follow the decomposition that coding agent itself gives,
a route through free space and a contact-rich behavior at its end.
First, for the route \circnum{1}, the model has no notion of a path
that clears the obstacle, and it takes the shortest one.
Second, it has no notion of replanning when the route it chose turns
infeasible midway.
Third, for the contact \circnum{2}, it does not recognize that the
contact itself carries a safety constraint.
A stronger planner does not close the gap (\cref{sec:ablation}), as
also reported at scale \citep{zhang2026despite}.
The safety constraint is simply never a core priority.

We propose \method to address the gap that the coding agent fails to accord safety the same priority as task completion. \method addresses this gap by equipping the model with two
obstacle-aware harnesses that give it the ability to prioritize the
safety constraint (\cref{fig:teaser}).
\emph{Obstacle-aware route planning}  (\cref{sec:route_planning}) handles \circnum{1}.
The objects in the scene are grounded as bounding boxes \citep{sam3},
and a route over them is drawn as a sequence of waypoints.
The agent plans the route ahead, verifies it, replans when a problem is
met along the way, and finally executes the path that has been cleared.
The safety constraint is thereby held at the same top priority as the
task goal.
\emph{Obstacle-aware contact execution} (\cref{sec:contact_execution}) handles \circnum{2}.
The contact strategy, meaning the exact contact position and whether
the gripper rotates, is decided by whether an obstacle stands next to
the contact location.
The same question is asked where the object is picked and where it is
placed.
Our core contributions are:
\begin{itemize}[leftmargin=1.4em, itemsep=1pt, topsep=2pt, parsep=0pt]
\item \textbf{Coding agents for safe robot control.} We study coding agent for robot control under a requirement it has never been scored on: finish the task under safety constraint. To our knowledge, this is the first coding agent for robot manipulation whose harness itself enforces safety: obstacle avoidance is verified throughout the task, including its contact-rich phases. It is also the first such agent evaluated on a collision-scored manipulation benchmark. 

\item \textbf{\method.} Our two obstacle-aware harnesses treat safety as a first-class objective, on equal footing with task completion. \circnum{1} Route planning plans, verifies, guards, and replans every free-space route. \circnum{2} Contact execution selects the contact position and wrist rotation from the spatial relation between the target and its neighboring obstacle.

\item \textbf{Results.} \method attains 81.2\% task success and 91.9\% collision avoidance on \bench , surpassing the previous SOTA by $13.7$ and $23.0$ points, respectively. Compared with the same agent without harnesses, these results represent gains of $31.2$ and $57.5$ points, respectively.

\end{itemize}

\section{Related Work}
\label{sec:related}

\paragraph{From VLAs and world-action models to coding agents.}
Vision-language-action (VLA) models map pixels and instructions
directly to actions.
RT-2 casts action prediction as next-token generation over a
vision-language backbone \citep{brohan2023rt2}, Octo trains a
generalist policy across many embodiments \citep{ghosh2024octo}, and
OpenVLA and its fine-tuning recipe scale an open backbone to new
manipulation tasks with modest data \citep{kim2024openvla,kim2025oft}.
The $\pi_0$ and $\pi_{0.5}$ family adds a flow-matching action head for
dexterous, contact-rich control \citep{black2024pi0,black2025pi05}.
A parallel line pretrains on video before it predicts action: UniPi
extracts actions from a generated goal video \citep{du2023unipi}, GR-1
and GR-2 predict future frames and actions jointly
\citep{wu2024gr1,cheang2024gr2}, and
WorldVLA and V-JEPA~2-AC couple world-model prediction and action
generation inside one architecture
\citep{cen2025worldvla,assran2025vjepa2}.
Both families are trained policies, so their competence reaches only
as far as their demonstrations.
A world-action model moreover inherits much of the brittleness of the
VLA it is built from \citep{zhang2026wamvsvla}, and in both cases the
decision resolved inside an action head resists inspection and
correction.
A coding agent instead has a language model write the controller as a
program that calls perception and control skills; written rather than
trained, such a policy transfers across tasks without robot-specific
data, can be read and edited before execution, and may compose
existing skills, learned policies included.

\paragraph{Coding agents in robot manipulation.}
A coding agent acts by writing and executing code rather than emitting
actions or language directly \citep{wang2024codeact,wang2023voyager}.
Code as Policies brought this idea to robot manipulation, with a
program composed from perception and control APIs serving as the
policy \citep{liang2023code}.
Follow-on work structured what the program could express, from the
prompt and its perception calls
\citep{singh2023progprompt,huang2023instruct2act} to object-centric
code with affordance and safety constraints \citep{mu2024robocodex}
and 3D value maps for a motion optimizer \citep{huang2023voxposer}.
A second line lets the agent improve its own code, rewriting it from
observed failures \citep{kumar2026actobserve}, growing a skill library
by evolutionary search \citep{lu2026aspire} or with a human in the loop
\citep{meng2025growing}, self-improving on real hardware
\citep{xiao2026enpire}, and trading test-time compute for reliability
\citep{fu2026capx}.
The strongest recent results come from GPT-6 Astra
\citep{openai2026astra}, which without robot-specific training reaches
near-perfect pick-and-place on a real arm
\citep{robocurve2026astra,jia2026agentpolicy} and, with a frozen
\pifive as a callable tool, handles bimanual, contact-rich tasks
\citep{su2026astra}.
The model itself emits no actions, so each of these results is as much
a statement about the harness as about the model.
A third line exposes a frozen policy as one callable primitive:
\harness identifies the scene once, crosses free space with analytic
inverse-kinematics primitives, and hands over to the frozen policy only
where the task turns contact-rich \citep{zhang2026harness}, and BATON
extends this design to long-horizon tasks \citep{xu2026baton}.
Across all three lines, the code plans and sequences skills, but a
skill executes as a straight-line motion or a learned primitive
regardless of what stands in its path; we are the first to study
coding agents for robot manipulation with safety placed on equal
footing with task success.

\paragraph{Safe control in robotics.}
Sampling- and optimization-based motion planners construct an explicit
route through free space and certify it against a collision model:
RRT grows a tree of feasible configurations toward the goal
\citep{lavalle1998rapidly}, and CHOMP and TrajOpt optimize a trajectory
against a swept-volume penalty
\citep{ratliff2009chomp,schulman2014motion}.
These planners fix a route before execution.
Runtime safety filters address that limitation: control barrier
functions wrap a controller in a filter that minimally edits each
commanded action to keep the system inside a safe set
\citep{ames2019cbf}, later extended to semantic, scene-grounded
regions \citep{brunke2025semantic}.
Such a filter edits whatever action it is handed; it cannot choose a
different grasp side or route.
\bench adds one obstacle, near the target or on the transport path, to
four tasks in each of the LIBERO-Spatial, Goal, Object, and Libero-10
suites \citep{liu2023libero}, scores an episode on task success and on
whether the obstacle ever moves, and pairs the benchmark with AEGIS, a
barrier-function layer over a frozen policy \citep{hu2025vlsa}.
LIBERO-Safety applies the same measurement to semantic constraints and
finds that scaling demonstrations tenfold barely lowers the collision
rate \citep{cui2026liberosafety}.
Across this line, safety is scored or filtered after a policy commits to a motion, rather than built into the decision that produces it. \method instead puts the check inside the decision, and the model that proposes the route could be replaced by any planner without changing the rest of the harness. we
are the first to bring the coding-agent paradigm to safe robot control.
\section{Method}
\label{sec:method}

\subsection{Preliminaries}
\label{sec:problem}

Coding agents have become a strong recipe for robot manipulation: a
language model writes the controller as a program, and the program
generalizes across tasks without robot-specific training.
Safe manipulation has so far been pursued along other routes: planners
that certify a trajectory in advance
\citep{lavalle1998rapidly,ratliff2009chomp,schulman2014motion}, filters
that edit each commanded action
\citep{ames2019cbf,brunke2025semantic,hu2025vlsa}, policies trained
under a safety objective \citep{zhang2025safevla,tang2026safedojo}.
Whether a coding agent can deliver safe manipulation has not been
asked, and we are the first to explore it.

\begin{figure}[t]
  \centering
  \includegraphics[width=0.99\linewidth]{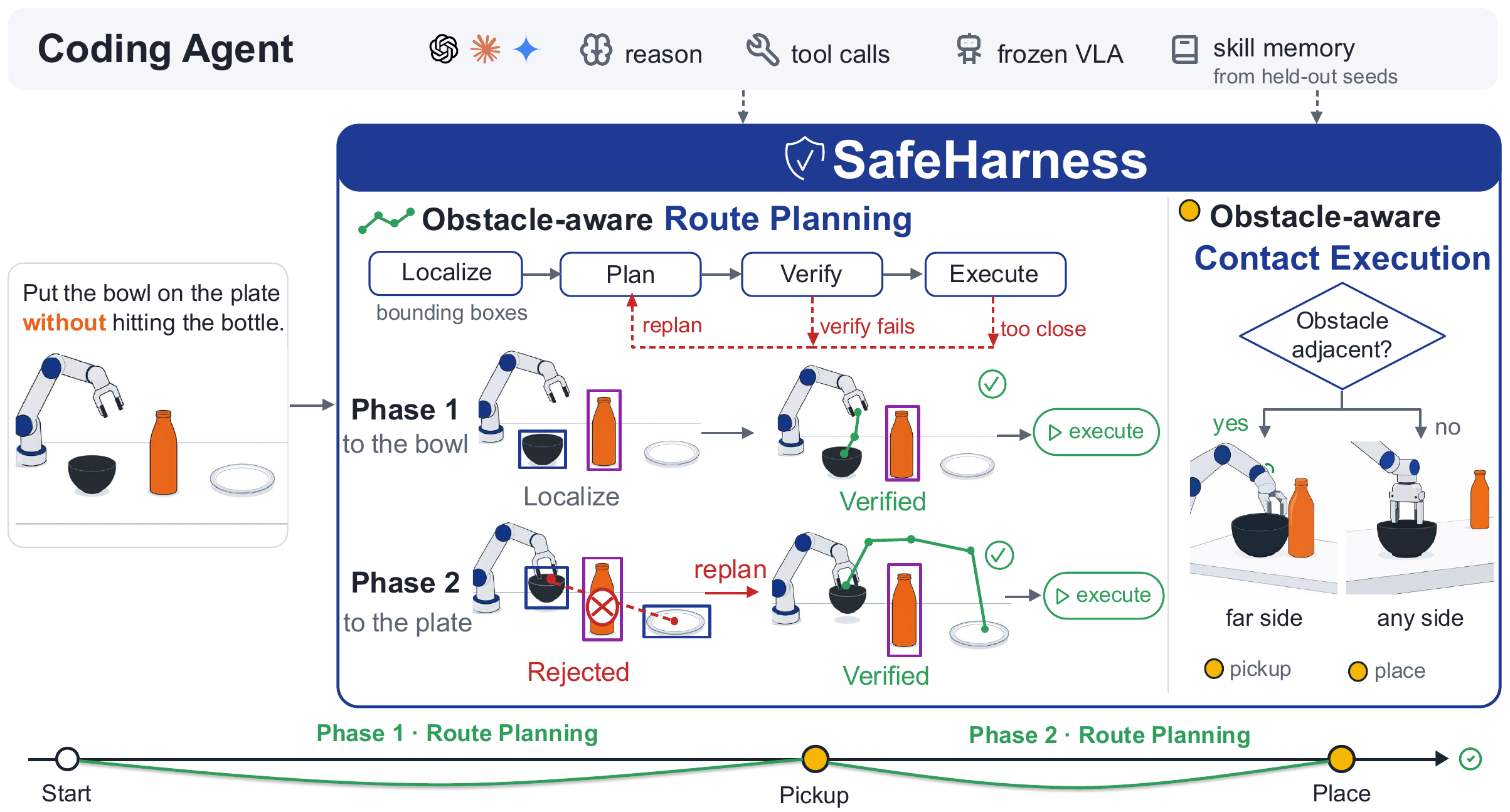}
  \vspace{-0.1in}
  \caption{\textbf{\method overview.}
  The task is cut at its contact events into phases, and both components
  run once per phase.
  \circnum{1} \emph{Obstacle-aware route planning}: the scene is
  grounded as boxes, the route of the current phase is planned,
  verified against the obstacle, and only then executed, with a failed
  verification or a halt during execution sending the loop back to
  planning.
  \circnum{2} \emph{Obstacle-aware contact execution}: the contact
  target is tested for an adjacent obstacle, and the contact is made
  from the far side whenever one is found.}
  \vspace{-0.2in}
  \label{fig:pipeline}
\end{figure}

We ask it here, and the recipe itself suggests how to approach it.
A coding agent composes a task from callable skills and motion
primitives, and the natural place to cut such a program is a contact
event.
Each phase is then a \emph{route} through free space followed by a
\emph{contact-rich behavior} at its end.
An obstacle can be touched in either
part.
The two parts nonetheless put it at risk in different ways.
A route sweeps a long path across the workspace, whereas a
contact-rich behavior sweeps a small volume that cannot be moved away
from the target it acts on.
We therefore split the safety constraint in the same place.
The route constraint asks for a clear path where no single point would collide with the obstacle.
The contact constraint asks that the contact-rich behavior be executed strategically when an obstacle is nearby.
Neither constraint implies the other, so an episode is safe only when
both of them hold.
The split gives \method its two components (\cref{fig:pipeline}).
\emph{Obstacle-aware route planning} (\cref{sec:route_planning}) is responsible
for the route constraint \circnum{1}.
\emph{Obstacle-aware contact execution} (\cref{sec:contact_execution}) is
responsible for the contact constraint \circnum{2}.
Both components run once per phase, so a task with two sub-goals
repeats them twice.

\subsection{Obstacle-Aware Route Planning}
\label{sec:route_planning}

A route is decided before the arm moves.
Planning and execution are asymmetric in cost: a plan can be revised
freely, whereas a motion already issued acts in the workspace and
cannot be recalled, so errors are best caught while they are still
words rather than movements.
\method accordingly plans the route of a phase ahead, verifies the
plan against the obstacle, and executes only a plan that has passed
verification.

Route planning begins anew with each phase rather than once for the
whole task.
Once the object is in the gripper, arm and object move as one body,
and the clearance a route requires changes with it.
A route cleared for the empty gripper is not cleared for the loaded one.
The first step of a phase is visual grounding.
The target and the obstacle are grounded as bounding boxes by a
segmentation model.
On that view the model is asked for a route as a polyline of
waypoints, under two conditions: it must end at the target, and it
must not cross the obstacle box.

The proposal obtained this way is not yet a route.
Before any motion is issued, it is checked against the obstacle box
outside the language model; the check covers the
gripper and, when an object is held, the object as well.
A candidate that intrudes into the box is rejected and the model
replans, until a route passes and is released for execution.

A route released for execution carries, however, only a limited
guarantee.
The verifier evaluates a single condition, namely whether the route
intersects the obstacle, and it does so on an approximately measured
scene.
Other physical constraints that arise during execution fall outside
its scope, and a route that passes verification may still fail
against them.
Replanning is therefore triggered at two points.
The first occurs at verification, before any motion is issued: if
the proposed route would enter the obstacle box, it is rejected and
a new route is planned.
The second occurs during execution, when the robot encounters
conditions the verifier did not evaluate.
The arm may approach the obstacle more closely than the plan
predicted, or it may become stuck partway along the route due to a
physical constraint that the bounding boxes did not capture.
In either case, execution halts and a new route is planned from the
pose at which the motion stopped.
Planning can thus be invoked repeatedly within a phase rather than
performed once at its outset.

Beyond replanning, execution imposes one further discipline: economy
of observation.
The natural impulse is to consult the route-plan image at every step,
yet each consultation appends the same view to the context window,
and a context window inflated by repeated identical observations
degrades the quality of the model's subsequent predictions.
\method therefore bounds the number of times the route-plan image may
be read during the execution of a route.
The bound removes redundant views without withholding information,
since a repeated image lengthens the context but adds nothing to
what the model knows.

Taken together, the phase runs as a plan--verify--execute cycle:
a route is proposed on the grounded scene, admitted only after it
clears the obstacle box, and executed under the same check, with
planning re-entered whenever verification or execution rejects the
current route.
What the cycle delivers is a path that has been confirmed clear
before and during motion, ending just above the contact target at a
pre-grasp or a release pose, where the phase is handed to the
contact step (\cref{sec:contact_execution}).
A long-horizon task is covered by repeating this cycle phase by
phase rather than by a longer plan.

\vspace{-0.07in}
\subsection{Obstacle-Aware Contact Execution}
\vspace{-0.1in}
\label{sec:contact_execution}

The second harness addresses the contact motion, the segment of the
phase that the route does not cover.
The route terminates above the contact target, at a pre-grasp or a
release pose; below this pose the gripper opens, descends, and
closes, and this motion sweeps a volume that route verification
does not examine.
Moreover, whereas the route is free to bend around the obstacle,
the contact motion is anchored to the target and cannot avoid its
vicinity.
When the obstacle is adjacent to the target, collisions therefore
occur in this final segment even when the route itself is
collision-free, and the contact motion must be planned with respect
to the obstacle as well.

The contact step is resolved in the same manner as the route: the
strategy is determined anew at each contact rather than fixed once
for the entire task.
Each contact begins with an adjacency test between the obstacle and
the contact target.
If no obstacle is adjacent, any contact strategy is admissible and
the step proceeds as it would in an uncluttered scene.
If an obstacle is adjacent, the contact is constrained to approach
from a direction that maintains clearance between the gripper and
the obstacle, and among the directions that satisfy this condition,
the direction farthest from the obstacle is selected.
If no direction satisfies the condition, the gripper is reoriented
so that its opening axis runs tangent to the obstacle, and the
payload is lowered vertically to avoid lateral motion
toward the obstacle.
The contact strategy is therefore governed by
obstacle clearance at every contact.

\subsection{Discussion: Constraints Under a Growing Context}

\label{sec:longhorizon}

The design of \method also addresses a property of coding agents that is independent of any particular constraint. An episode unfolds as a single conversation spanning hundreds of steps, in which every tool call appends its result and every observation appends an image. By the time the arm approaches the obstacle, the context holds tens of thousands of tokens, while the constraint is stated near its beginning. Language models exploit such content less reliably as the context grows: recall degrades for material in the middle of long inputs \citep{liu2023lostmiddle}, instruction following deteriorates with length \citep{wu2025lifbench}, system-prompt adherence drifts within a few turns \citep{li2024instruction}, performance declines over multi-turn conversations \citep{laban2025lost}, and even trained safety behavior weakens under sufficiently long contexts \citep{anil2024manyshot}. A constraint conveyed solely through the prompt is thus exposed to the very context that task execution produces, and a stronger backbone does not remove this exposure (\cref{sec:ablation}). \method does not rely on the constraint surviving the context. The route is planned and verified before the arm moves, using tools that are stateless with respect to the conversation: they maintain the obstacle box outside the context and return results on which the planner acts in the next step. Safety is thereby enforced explicitly wherever a motion is decided, rather than requested once at the outset. 
\section{Experiments}

\subsection{Experimental Setup}

\label{sec:exp}


\noindent\textbf{Benchmark.} \citep{hu2025vlsa} is constructed from four representative tasks drawn from each of the LIBERO-Spatial, Goal, Object, and Libero-10 suites \citep{liu2023libero} yielding 16 base tasks.
Each augmented with one obstacle object selected from six household items: a moka pot, storage box, milk carton, wine bottle, mug, or book. Each task is instantiated at two difficulty levels: in Level I, the obstacle is placed in close proximity to the target object, whereas in Level II it is placed away from the target but directly on the natural transport path, resulting in 32 tasks in total. 
We evaluate \method on all 32 \bench tasks with 10 seeds per task, and on real-robot using PiPER-X 6-DoF research arm for two tasks with 20 trials each.

\label{sec:setup}

\noindent\textbf{Metrics.}
Following the official SafeLIBERO evaluator, an obstacle is considered displaced if it
moves by more than 1\,mm at any point during an episode.
We report three metrics: task success rate (TSR), the percentage of episodes that achieve
the task goal; collision-avoidance rate (CAR), the percentage of episodes in which no
obstacle is displaced; and safe success rate (SSR), the percentage of episodes that
achieves the task goal with no displaced obstacle.

\noindent\textbf{Models.}
Our system uses a coding agent to compose a frozen VLA policy with the obstacle-aware harnesses introduced in \cref{sec:method}. We use the
same frozen VLA policy as \harness \citep{zhang2026harness} without any fine-tuning, SAM-3 \citep{sam3} for segmentation, and either GPT-5.5 or GPT-6-Astra \citep{openai2026astra} as the coding-agent backbone,
reporting results for both backbones in \cref{tab:main} -- \ref{tab:ablation}. To keep comparisons controlled, the coding-agent loop and the LIBERO skill memory \citep{liu2023libero} follow the configuration of \harness \citep{zhang2026harness}, so that performance differences can be attributed to the proposed harnesses rather than to the surrounding infrastructure. 
In the ablation study, the backbone alone, without tools, skills, or memory, serves as the \emph{Model} variant, and this shared configuration without our harnesses serves as \emph{Model + Skills} (\cref{sec:ablation}).
In the main experiments, we compare against two foundation VLA models, \pifive \citep{black2025pi05} and OpenVLA-OFT \citep{kim2025oft}, as well as the previous state-of-the-art method AEGIS \citep{hu2025vlsa}, which augments \pifive with a barrier-function-based safety layer.

\noindent\textbf{Implementation Details.}
The agent receives the arm's proprioceptive state at every step and consults
the third-person and wrist views on demand.
A route is considered to collide with the obstacle if the gripper (or, when
loaded, the held object) passes within 12\,cm of it or less than 5\,cm above it.
The verifier rejects any such route, and every motion segment is subjected to
the same check before execution. If the gripper stalls more than 3\,cm short of
a waypoint, the route is replanned from the current pose. The adjacency test is
triggered when the gripper footprint at the contact target comes within 2\,cm of
the obstacle, and the VLA policy is invoked only once the gripper is within
3\,cm of the target. 
On \bench, an episode is allowed at most 900 environment steps; a successful run takes about 400
steps on average, and about 700 on Libero-10, whose tasks comprise two sub-goals.

\subsection{Main Results}
\label{sec:main}

The source of these gains lies in how each method accounts for
obstacles during execution. A frozen VLA policy never encounters
obstacle avoidance in its training objective, so nothing inside the
policy distinguishes an approach that clears the obstacle from one that
sweeps through it. A barrier layer such as AEGIS sits downstream of the
policy and edits the commanded action: it can veto a motion that
violates the safety constraint but has no means of proposing a better
one, so each intervention halts an unsafe motion without steering the
policy back toward the goal, and this cost compounds when interventions
recur. On the Libero-10 suite, whose two sub-goals each call for such an
intervention, AEGIS even falls below the frozen policy it wraps in task
success. The vanilla GPT-5.5 agent can in principle reason about
obstacles, but its awareness of safety derives entirely from the prompt,
so its safety behavior depends on whether the constraint remains salient
as the context grows over the episode, which we examine in
\cref{sec:ablation}. \method, in contrast, grounds the obstacles in the
scene and plans the route and execution steps around them before acting,
so that collision avoidance is built into the plan rather than enforced
after the fact or left to the prompt. This design retains the
adaptability of a coding agent while avoiding the fixed behavior of
VLA policies that replay patterns learned from training data.

The real-robot results in \cref{tab:real_exp} follow the same pattern, with \method{} leading on both tasks under both metrics. We compare against \pifive{} and a vanilla coding agent built on GPT-5.5, which shares the backbone of \method{} but lacks our harness. \pifive{} performs worst and rarely avoids collisions. Its trajectory remains largely unchanged whether the obstacle sits beside the target or lies across the path, which reflects the insensitivity to scene layout discussed above. The vanilla agent adapts more readily, varying its approach routes and retrying after failures, yet it still collides frequently. Its traces show explicit reasoning about obstacles, but this reasoning is not translated into the plan it generates (\cref{sec:whyfail}). The result is a persistent gap between what the agent recognizes and what it executes. \method{} stays robust under both obstacle placements, maintaining high task success while seldom colliding, which indicates that safety and task completion need not be traded against each other. On physical hardware, this robustness again follows from resolving obstacles at the planning stage. Once the route is committed with the obstacle accounted for, every subsequent step inherits the constraint rather than depending on the agent to keep it in view.


\begin{table}[t]
\caption{\bench \citep{hu2025vlsa}: per-suite and average \tsr and \car,
in \%. \method instantiated with two coding-agent backbones (GPT-5.5 and GPT-6-Astra). Best per column in bold.}
\label{tab:main}
\vspace{-0.1in}
\centering
\small
\renewcommand{\arraystretch}{0.92}
\setlength{\tabcolsep}{4pt}
\begin{tabular}{l cc cc cc cc cc}
\toprule
& \multicolumn{2}{c}{Spatial} & \multicolumn{2}{c}{Goal} & \multicolumn{2}{c}{Object} & \multicolumn{2}{c}{Libero-10} & \multicolumn{2}{c}{Average} \\
\cmidrule(lr){2-3}\cmidrule(lr){4-5}\cmidrule(lr){6-7}\cmidrule(lr){8-9}\cmidrule(lr){10-11}
Method & \tsr & \car & \tsr & \car & \tsr & \car & \tsr & \car & \tsr & \car \\
\midrule
OpenVLA-OFT \citep{kim2025oft} & 36.5 & 8.3 & 24.3 & 19.5 & 28.8 & 11.5 & 15.3 & 6.0 & 26.2 & 11.3 \\
\pifive \citep{black2025pi05} & 59.3 & 14.0 & 60.0 & 20.0 & 57.5 & 18.0 & 54.3 & 16.5 & 57.8 & 17.1 \\
AEGIS \citep{hu2025vlsa} & 68.5 & 68.0 & \textbf{82.8} & 76.5 & 72.5 & 71.3 & 46.3 & 59.8 & 67.5 & 68.9 \\
\midrule
\rowcolor{blue!8} \method (GPT-5.5) & \textbf{81.2} & 76.2 & 68.8 & 88.8 & 90.0 & 87.5 & 75.0 & 95.0 & 78.8 & 86.9 \\
\rowcolor{blue!15} \method (GPT-6-Astra) & 76.2 & \textbf{90.0} & 78.8 & \textbf{90.0} & \textbf{92.5} & \textbf{91.2} & \textbf{77.5} & \textbf{96.2} & \textbf{81.2} & \textbf{91.9} \\
\bottomrule
\end{tabular}
\vspace{-0.13in}
\end{table}

\begin{table}[t]
\caption{Real-robot results, \tsr / \car in \%, 20 trials per task. Each
scene holds two obstacles, one beside the target object and one on the
path to the plate. Small numbers give the relative improvement of \method{}
over the GPT-5.5 coding agent with the same backbone. Best per column in bold.}
\label{tab:real_exp}
\vspace{-0.13in}
\centering
\small
\renewcommand{\arraystretch}{0.92}
\setlength{\tabcolsep}{7pt}
\begin{tabular}{l cc cc cc}
\toprule
& \multicolumn{2}{c}{White can $\to$ plate} & \multicolumn{2}{c}{Coke bottle $\to$ plate} & \multicolumn{2}{c}{Average} \\
\cmidrule(lr){2-3}\cmidrule(lr){4-5}\cmidrule(lr){6-7}
Method & \tsr & \car & \tsr & \car & \tsr & \car \\
\midrule
\pifive & 20 & 15 & 30 & 20 & 25.0 & 17.5 \\
GPT-5.5 & 30 & 40 & 25 & 50 & 27.5 & 45.0 \\
\rowcolor{blue!15}\method (GPT-5.5) & \imp{70}{133} & \imp{85}{113} & \imp{60}{140} & \imp{80}{60} & \imp{65.0}{136} & \imp{82.5}{83} \\
\bottomrule
\end{tabular}
\vspace{-0.1in}
\end{table}


\vspace{-0.07in}
\subsection{Ablation: Where the Coding Agent Fails, and What Fixes It}
\label{sec:ablation}
\label{sec:whyfail}

We fix the 32 scenes and the ten seeds per task, so that every row rests
on 160 episodes per level, and cross two backbones with the three settings of
\cref{sec:setup}. \emph{Model} measures the intrinsic capability of the
backbone, and \emph{Model + Skills} tests whether an agent that already
completes the tasks can respect a constraint given only as language.

As shown in \cref{tab:ablation}, a coding agent does not reliably act on
a safety constraint it has understood, and adding task skills can widen
rather than close this gap. Without external tools, \emph{Model} fails
on both metrics, since the backbone alone lacks the execution abilities
manipulation requires. Its \tsr is lowest at Level~II, where transport
must be planned from scratch, and its higher \car there comes from
episodes that fail before reaching the obstacle, so safe successes stay
rare. Skills raise \tsr as expected, yet do not
make execution safer: at Level~I, \car and \ssr fall below those of
\emph{Model} for both backbones. When a stored skill matches the subtask, the agent replays
it instead of reconciling it with the constraint, and since these skills
were acquired in obstacle-free scenes, replaying them near the target
leads to collisions. Transport is rarely covered by a stored skill, so
the agent plans that motion from scratch, and mainly then does the
constraint enter its reasoning. A constraint that exists only as
language is thus honored mostly when no ready-made solution is at hand,
so that competence and safety compete rather than reinforce each other;
a stronger backbone does not change this, as \emph{Model + Skills} with
GPT-6-Astra is the least safe setting in the table.

\method resolves this tension by making safety part of the planning
procedure rather than a sentence in the prompt. Because the agent must
ground the obstacles, plan the route, and settle the contact strategy
before acting, the constraint shapes execution whether or not a matching
skill exists. \method thus raises \tsr and \car together,
improving on \emph{Model + Skills} by $22.6$ and $43.1$ points with
GPT-5.5 and by $31.2$ and $57.5$ points with GPT-6-Astra, and \ssr more than
doubles. Since both share backbone, tools, and skill memory, the gains
isolate the harnesses.

\begin{table}[t]
\vspace{-0.1in}
\caption{Ablation on model family and the source of safety gains. \tsr / \car / \ssr in \%.
\emph{\textbf{Model}}: the language model drives motion and primitives directly.
\emph{\textbf{Model + Skills}}: adds skill memory and the frozen VLA.
\emph{\method}: further adds the obstacle-aware harnesses of \cref{sec:method}.
Small numbers give the relative improvement of \method{} over Model + Skills with the same planner.}
\vspace{-0.1in}
\label{tab:ablation}
\centering
\small
\renewcommand{\arraystretch}{0.98}
\setlength{\tabcolsep}{3.5pt}
\scalebox{0.95}{
\begin{tabular}{l l ccc ccc ccc}
\toprule
& & \multicolumn{3}{c}{Level~I} & \multicolumn{3}{c}{Level~II} & \multicolumn{3}{c}{All 32} \\
\cmidrule(lr){3-5}\cmidrule(lr){6-8}\cmidrule(lr){9-11}
Planner & Agent & \tsr & \car & \ssr & \tsr & \car & \ssr & \tsr & \car & \ssr \\
\midrule
\multirow{3}{*}{GPT-5.5} & Model & 43.1 & 43.1 & 25.6 & 19.4 & 56.9 & 18.8 & 31.3 & 50.0 & 22.2 \\
& Model + Skills & 44.4 & 18.1 & 13.1 & 68.1 & 69.4 & 49.4 & 56.2 & 43.8 & 31.3 \\
 & \cellcolor{blue!15}\method & \cellcolor{blue!15}\imp{86.2}{94} & \cellcolor{blue!15}\imp{90.0}{397} & \cellcolor{blue!15}\imp{80.0}{511} & \cellcolor{blue!15}\imp{71.2}{5} & \cellcolor{blue!15}\imp{83.8}{21} & \cellcolor{blue!15}\imp{66.3}{34} & \cellcolor{blue!15}\imp{78.8}{40} & \cellcolor{blue!15}\imp{86.9}{98} & \cellcolor{blue!15}\imp{73.1}{134} \\
\midrule
\multirow{3}{*}{GPT-6-Astra} & Model & 36.9 & 50.6 & 30.6 & 31.9 & 74.4 & 31.3 & 34.4 & 62.5 & 30.9 \\
 & Model + Skills & 56.9 & 19.4 & 11.9 & 43.1 & 49.4 & 25.6 & 50.0 & 34.4 & 18.8 \\
 & \cellcolor{blue!15}\method & \cellcolor{blue!15}\imp{90.0}{58} & \cellcolor{blue!15}\imp{92.5}{377} & \cellcolor{blue!15}\imp{83.8}{604} & \cellcolor{blue!15}\imp{72.5}{68} & \cellcolor{blue!15}\imp{91.2}{85} & \cellcolor{blue!15}\imp{72.5}{183} & \cellcolor{blue!15}\imp{81.2}{62} & \cellcolor{blue!15}\imp{91.9}{167} & \cellcolor{blue!15}\imp{78.1}{315} \\
\bottomrule
\end{tabular}}
\end{table}

\begin{table}[t]
\caption{Component ablation on \bench, \tsr / \car / \ssr in \%. Rows
add one harness at a time to \emph{Model + Skills} (MS). 
Small numbers give the relative
improvement over \emph{MS + Route}.}
\vspace{-0.1in}
\label{tab:component}
\centering
\small
\renewcommand{\arraystretch}{0.92}
\setlength{\tabcolsep}{3.5pt}
\scalebox{0.95}{
\begin{tabular}{l l ccc ccc ccc}
\toprule
& & \multicolumn{3}{c}{Level~I} & \multicolumn{3}{c}{Level~II} & \multicolumn{3}{c}{All 32} \\
\cmidrule(lr){3-5}\cmidrule(lr){6-8}\cmidrule(lr){9-11}
Planner & Agent & \tsr & \car & \ssr & \tsr & \car & \ssr & \tsr & \car & \ssr \\
\midrule
\multirow{4}{*}{GPT-5.5}
 & Model + Skills (MS) & 44.4 & 18.1 & 13.1 & 68.1 & 69.4 & 49.4 & 56.2 & 43.8 & 31.3 \\
 & MS + Contact & 55.0 & 38.8 & 32.5 & 68.8 & 70.0 & 50.0 & 61.9 & 54.4 & 41.3 \\
 & MS + Route   & 75.0 & 71.3 & 62.5 & 71.2 & 82.5 & 58.8 & 73.1 & 76.9 & 60.6 \\
 & \cellcolor{blue!15}\method & \cellcolor{blue!15}\imp{86.2}{15} & \cellcolor{blue!15}\imp{90.0}{26} & \cellcolor{blue!15}\imp{80.0}{28} & \cellcolor{blue!15}\imp{71.2}{0} & \cellcolor{blue!15}\imp{83.8}{2} & \cellcolor{blue!15}\imp{66.3}{13} & \cellcolor{blue!15}\imp{78.8}{8} & \cellcolor{blue!15}\imp{86.9}{13} & \cellcolor{blue!15}\imp{73.1}{21} \\
\midrule
\multirow{4}{*}{GPT-6-Astra}
 & Model + Skills (MS) & 56.9 & 19.4 & 11.9 & 43.1 & 49.4 & 25.6 & 50.0 & 34.4 & 18.8 \\
 & MS + Contact Harness & 67.5 & 45.0 & 40.0 & 46.3 & 53.8 & 38.8 & 56.9 & 49.4 & 39.4 \\
 & MS + Route Harness  & 78.8 & 72.5 & 65.0 & 70.0 & 87.5 & 63.8 & 74.4 & 80.0 & 64.4 \\
 & \cellcolor{blue!15}\method & \cellcolor{blue!15}\imp{90.0}{14} & \cellcolor{blue!15}\imp{92.5}{28} & \cellcolor{blue!15}\imp{83.8}{29} & \cellcolor{blue!15}\imp{72.5}{4} & \cellcolor{blue!15}\imp{91.2}{4} & \cellcolor{blue!15}\imp{72.5}{14} & \cellcolor{blue!15}\imp{81.2}{9} & \cellcolor{blue!15}\imp{91.9}{15} & \cellcolor{blue!15}\imp{78.1}{21} \\
\bottomrule
\end{tabular}}
\vspace{-0.1in}
\end{table}

\subsection{Ablation: Route Planning versus Contact Execution}

\label{sec:component}

\cref{sec:ablation} evaluates the harnesses as a whole; here we ablate the two components of the safety harness, route planning (\cref{sec:route_planning}) and contact execution (\cref{sec:contact_execution}), to quantify the contribution of each. When a component is removed, the corresponding phase is left to the agent itself: \emph{MS + Route} lets the agent perform grasp and release on its own, and \emph{MS + Contact} lets it plan its own transport motions. Two findings emerge. First, both components are necessary. Removing either one lowers \ssr across all base models and both levels, and neither partial configuration reaches the full method, indicating that the two address distinct rather than redundant failure modes. Second, route planning contributes more than contact execution. At the moment of contact, the agent's own grasps typically follow a standard approach direction and pose that often keep the gripper clear of a nearby obstacle, so many contact-phase hazards are avoided without the harness. During transport, however, nothing in the agent's own motion inherently avoids an obstacle on the path, so the obstacle is struck unless the trajectory is explicitly rerouted. This effect extends to Level~I as well: even when the obstacle stands beside the target, a tall object such as a wine bottle sweeps the surrounding space as soon as it is lifted, so the motion immediately after grasping already falls within the scope of route planning. The two failure modes of \cref{sec:method} are therefore distinct, and each harness component removes one of them.

\subsection{Qualitative Analysis}

\label{sec:qual}

\begin{wrapfigure}{r}{0.6\linewidth}
  \centering
  \vspace{-0.2in}
  \includegraphics[width=\linewidth]{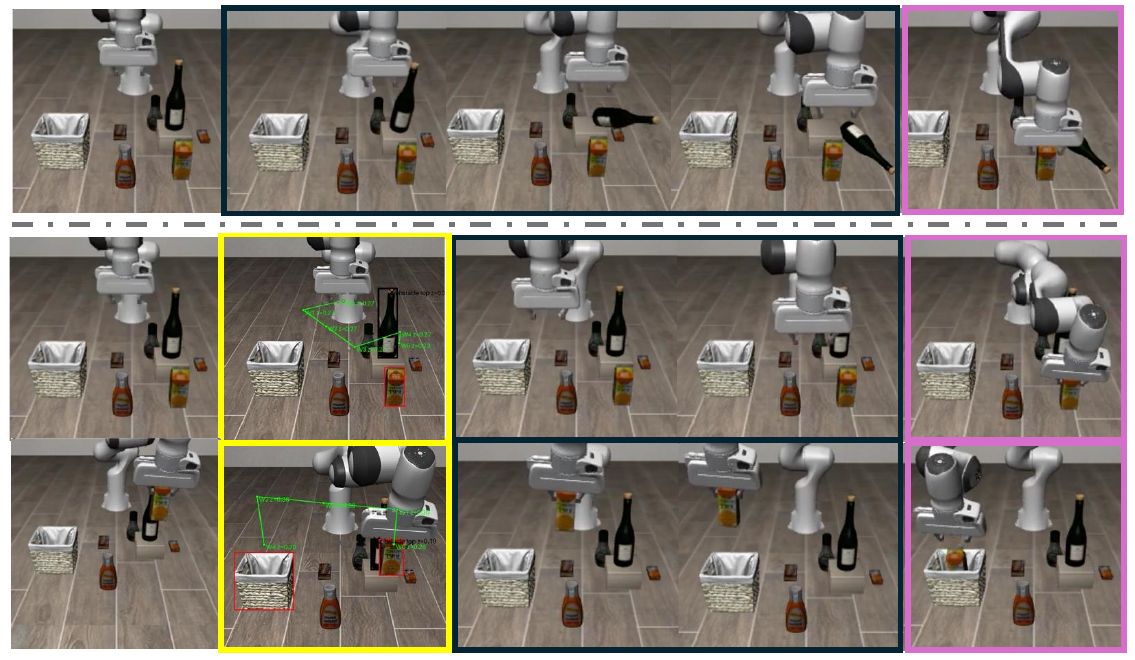}
  \caption{\textbf{Rollouts on \bench with and without \method.}
  The task is to place the orange juice in the basket without hitting the wine bottle. Without \method (top row), 
  hits the obstacle en route to the grasp; with \method (bottom rows), the agent detours around it and finishes collision-free. Panel frames: yellow = route planning (bounding boxes and waypoints), black = verified-route execution, purple = contact-rich moments.}
  \label{fig:qual}
  \vspace{-0.2in}
\end{wrapfigure}

\cref{fig:qual} contrasts a failed and a successful rollout on the same \bench \citep{hu2025vlsa} scene, where the orange juice must reach the basket without touching the wine bottle.The skills-only agent (top row) acknowledges the bottle in its reasoning, but the acknowledgment never shapes the
plan that follows, which takes the shortest line at grasp height, straight through the bottle, as if success were
defined without the constraint rather than under it.
\begin{wrapfigure}{r}{0.55\textwidth}
    \centering
    \vspace{-0.15in}
    \includegraphics[width=\linewidth]{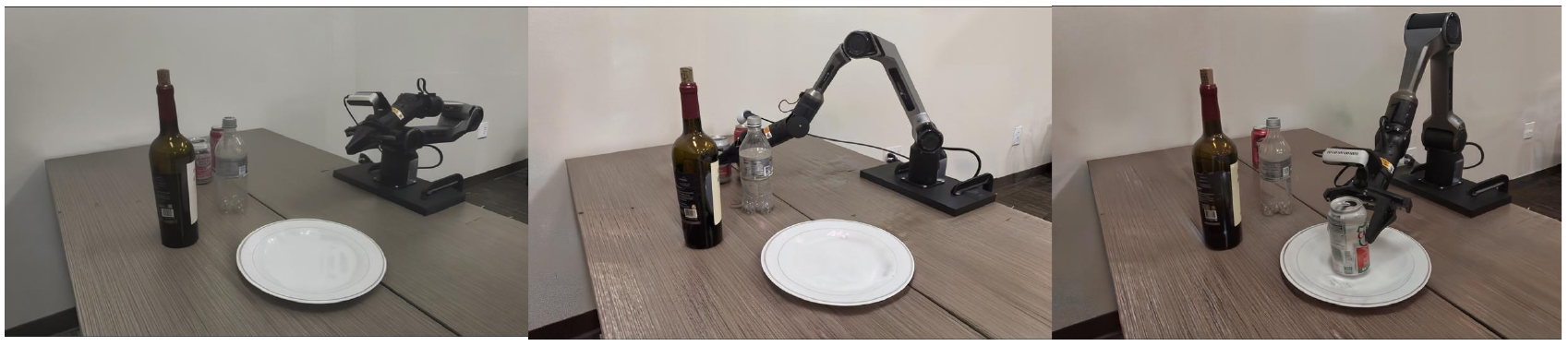}
    \caption{\textbf{Real robot experiment demo}. \method{} leads to safe and successful task completion.}
    \label{fig:real_robot}
    \vspace{-0.2in}
\end{wrapfigure}
 \method (bottom rows) instead
  clears the bottle and completes the task, because every route is checked before the arm moves rather than left to the
  model's reasoning. On Libero-10 this loop runs twice per episode, once per sub-goal, and the second leg re-uses the obstacle box measured for the first.
\cref{fig:real_robot} shows a successful rollout on the real robot.

We now walk through the successful rollout phase by phase, each running the full loop of \cref{sec:route_planning} and \cref{sec:contact_execution}. In the first phase the agent grounds the juice and the bottle as boxes and plans the route to the pick; the bottle blocks the way, so the route detours around it, passes verification, and execution follows the verified plan. In the second phase, with the bottle no longer in the way, the route to the basket proceeds directly and is verified as proposed, so the harness plans a detour only when an obstacle demands one and safety is not bought by timidity. At both contact moments, the adjacency test finds no obstacle beside the target, so contact execution admits the default strategy. The rollout thus exercises every decision point of \method, each intervening exactly where the scene requires it.

\vspace{-0.1in}
\section{Conclusions}
\label{sec:conclusion}
\vspace{-0.1in}

Safety remains a critical yet largely overlooked dimension of coding agents for robot manipulation, whose progress has thus far been measured solely by task completion. To address this gap, we present \method{}, the first coding-agent framework designed for safe robot control. \method{} comprises two complementary components: obstacle-aware route planning, which plans routes in advance and replans them once they become infeasible, and obstacle-aware contact execution, which keeps contact-rich motion clear of surrounding obstacles. Together, they elevate safety to the same priority as task completion. On \bench{}, \method{} surpasses the previous state of the art by $13.7$ points in task success and $23.0$ points in collision avoidance. Compared with the same agent without
harnesses, these results represent gains of 31.2 and 57.5 points, respectively.
More broadly, our findings suggest that constraints that must hold throughout execution cannot be entrusted to the agent's reasoning alone, but should instead be embedded in the mechanisms through which it acts.

\bibliography{references}

@inproceedings{liang2023code,
  title     = {Code as Policies: Language Model Programs for Embodied Control},
  author    = {Liang, Jacky and Huang, Wenlong and Xia, Fei and Xu, Peng and
               Hausman, Karol and Ichter, Brian and Florence, Pete and Zeng, Andy},
  booktitle = {IEEE International Conference on Robotics and Automation (ICRA)},
  year      = {2023},
  note      = {arXiv:2209.07753}
}

@inproceedings{singh2023progprompt,
  title     = {ProgPrompt: Generating Situated Robot Task Plans using Large Language Models},
  author    = {Singh, Ishika and Blukis, Valts and Mousavian, Arsalan and Goyal, Ankit and
               Xu, Danfei and Tremblay, Jonathan and Fox, Dieter and Thomason, Jesse and
               Garg, Animesh},
  booktitle = {IEEE International Conference on Robotics and Automation (ICRA)},
  year      = {2023},
  note      = {arXiv:2209.11302}
}

@article{huang2023instruct2act,
  title   = {Instruct2Act: Mapping Multi-modality Instructions to Robotic Actions with Large Language Model},
  author  = {Huang, Siyuan and Jiang, Zhengkai and Dong, Hao and Qiao, Yu and Gao, Peng and Li, Hongsheng},
  journal = {arXiv preprint arXiv:2305.11176},
  year    = {2023}
}

@article{huang2023voxposer,
  title     = {VoxPoser: Composable 3D Value Maps for Robotic Manipulation with Language Models},
  author    = {Huang, Wenlong and Wang, Chen and Zhang, Ruohan and Li, Yunzhu and
               Wu, Jiajun and Fei-Fei, Li},
  journal={arXiv preprint arXiv:2307.05973},
  year      = {2023},
}

@article{mu2024robocodex,
  title     = {RoboCodeX: Multimodal Code Generation for Robotic Behavior Synthesis},
  author    = {Mu, Yao and Chen, Junting and Zhang, Qinglong and Chen, Shoufa and Yu, Qiaojun and Ge, Chongjian and Chen, Runjian and Liang, Zhixuan and Hu, Mengkang and Tao, Chaofan and others},
  journal={arXiv preprint arXiv:2402.16117},
  year      = {2024},
}

@article{zhang2026despite,
  title   = {Using Large Language Models for Embodied Planning Introduces Systematic Safety Risks},
  author  = {Zhang, Tao and Qu, Kaixian and Li, Zhibin and Wu, Jiajun and Hutter, Marco and
             Li, Manling and Shi, Fan},
  journal = {arXiv preprint arXiv:2604.18463},
  year    = {2026}
}

@inproceedings{liu2023libero,
  title     = {LIBERO: Benchmarking Knowledge Transfer for Lifelong Robot Learning},
  author    = {Liu, Bo and Zhu, Yifeng and Gao, Chongkai and Feng, Yihao and Liu, Qiang and
               Zhu, Yuke and Stone, Peter},
  booktitle = {Advances in Neural Information Processing Systems},
  year      = {2023},
}

@article{hu2025vlsa,
  title   = {VLSA: Vision-Language-Action Models with Plug-and-Play Safety Constraint Layer},
  author  = {Hu, Songqiao and Liu, Zeyi and Liu, Shuang and Cen, Jun and Meng, Zihan and
             Wang, Shihefeng and Li, Xiang and He, Xiao},
  journal={arXiv preprint arXiv:2512.11891},
  year={2025}
}

@article{black2024pi0,
  title   = {$\pi_0$: A Vision-Language-Action Flow Model for General Robot Control},
  author  = {Black, Kevin and Brown, Noah and Driess, Danny and Esmail, Adnan and Equi, Michael and Finn, Chelsea and Fusai, Niccolo and Groom, Lachy and Hausman, Karol and Ichter, Brian and others},
  journal = {arXiv preprint arXiv:2410.24164},
  year    = {2024}
}

@article{black2025pi05,
  title   = {$\pi_{0.5}$: a Vision-Language-Action Model with Open-World Generalization},
  author  = {{Physical Intelligence} and Black, Kevin and Brown, Noah and Darpinian, James and Dhabalia, Karan and Driess, Danny and Esmail, Adnan and Equi, Michael and Finn, Chelsea and Fusai, Niccolo and others},
  journal = {arXiv preprint arXiv:2504.16054},
  year    = {2025}
}

@article{kim2024openvla,
  title   = {OpenVLA: An Open-Source Vision-Language-Action Model},
  author  = {Kim, Moo Jin and Pertsch, Karl and Karamcheti, Siddharth and Xiao, Ted and Balakrishna, Ashwin and Nair, Suraj and Rafailov, Rafael and Foster, Ethan and Lam, Grace and Sanketi, Pannag and others},
  journal = {arXiv preprint arXiv:2406.09246},
  year    = {2024}
}

@article{kim2025oft,
  title     = {Fine-Tuning Vision-Language-Action Models: Optimizing Speed and Success},
  author    = {Kim, Moo Jin and Finn, Chelsea and Liang, Percy},
  journal={arXiv preprint arXiv:2502.19645},
  year      = {2025},
}

@article{brohan2023rt2,
  title   = {RT-2: Vision-Language-Action Models Transfer Web Knowledge to Robotic Control},
  author  = {Brohan, Anthony and Brown, Noah and Carbajal, Justice and Chebotar, Yevgen and Chen, Xi and Choromanski, Krzysztof and Ding, Tianli and Driess, Danny and Dubey, Avinava and Finn, Chelsea and others},
  journal = {arXiv preprint arXiv:2307.15818},
  year    = {2023}
}

@inproceedings{ghosh2024octo,
  title   = {Octo: An Open-Source Generalist Robot Policy},
  author  = {{Octo Model Team} and Ghosh, Dibya and Walke, Homer and Pertsch, Karl and Black, Kevin and Mees, Oier and Dasari, Sudeep and Hejna, Joey and Kreiman, Tobias and Xu, Charles and others},
  booktitle = {Robotics: Science and Systems (RSS)},
  year    = {2024},
  note    = {arXiv:2405.12213}
}

@inproceedings{zhang2025safevla,
  title     = {SafeVLA: Towards Safety Alignment of Vision-Language-Action Model via Constrained Learning},
  author    = {Zhang, Borong and Zhang, Yuhao and Ji, Jiaming and Lei, Yingshan and Cai, Yishuai and
               Dai, Josef and Chen, Yuanpei and Yang, Yaodong},
  booktitle = {Advances in Neural Information Processing Systems (NeurIPS)},
  year      = {2025},
  note      = {Spotlight; arXiv:2503.03480}
}

@article{tang2026safedojo,
  title   = {SafeDojo: Safe Reinforcement Learning for VLA via Interactive World Model},
  author  = {Tang, Kai and Jia, Peidong and Chu, Zhong and Wu, Jixian and Ma, Rui and Cao, Jiajun and Zhao, Fangyuan and Chen, Sixiang and Guo, Yichen and Chi, Xiaowei and others},
  journal = {arXiv preprint arXiv:2606.20698},
  year    = {2026}
}

@inproceedings{cui2026liberosafety,
  title   = {LIBERO-Safety: A Comprehensive Benchmark for Physical and Semantic Safety in Vision-Language-Action Models},
  author  = {Cui, Rongxu and Zhang, Zongzheng and Pang, Jingrui and Chi, Haohan and Guo, Jinbang and Zhang, Saining and Xie, Shaoxuan and Jin, Xin and Mu, Yao and Yang, Jiaolong and others},
  booktitle={European Conference on Computer Vision},
  year={2026},
  organization={Springer}
}

@article{huang2026safemanip,
  title   = {SafeManip: A Property-Driven Benchmark for Temporal Safety Evaluation in Robotic Manipulation},
  author  = {Huang, Chengyue and Huynh, Khang Vo and Elbaum, Sebastian and Kira, Zsolt and Feng, Lu},
  journal = {arXiv preprint arXiv:2605.12386},
  year    = {2026}
}

@article{fan2026safevlabench,
  title   = {SafeVLA-Bench: A Benchmark for the Success-Safety Gap in Vision-Language-Action Models},
  author  = {Fan, Jialiang and Xu, Weizhe and Sokolsky, Oleg and Lee, Insup and Kong, Fanxin},
  journal = {arXiv preprint arXiv:2606.00773},
  year    = {2026}
}

@article{lavalle1998rapidly,
  title       = {Rapidly-exploring Random Trees: A New Tool for Path Planning},
  author      = {LaValle, Steven M.},
  journal={Research Report 9811},
  year={1998},
  publisher={Department of Computer Science, Iowa State University}
}

@inproceedings{ratliff2009chomp,
  title     = {CHOMP: Gradient Optimization Techniques for Efficient Motion Planning},
  author    = {Ratliff, Nathan and Zucker, Matthew and Bagnell, J. Andrew and Srinivasa, Siddhartha},
  booktitle = {IEEE International Conference on Robotics and Automation (ICRA)},
  year      = {2009}
}

@article{schulman2014motion,
  title   = {Motion Planning with Sequential Convex Optimization and Convex Collision Checking},
  author  = {Schulman, John and Duan, Yan and Ho, Jonathan and Lee, Alex and Awwal, Ibrahim and
             Bradlow, Henry and Pan, Jia and Patil, Sachin and Goldberg, Ken and Abbeel, Pieter},
  journal = {The International Journal of Robotics Research},
  year    = {2014},
}

@article{zhang2026harness,
  title={Harness VLA: Steering Frozen VLAs into Reliable Manipulation Primitives via Memory-Guided Agents},
  author={Zhang, Yixian and Zhang, Huanming and Gao, Feng and Li, Xiao and Liu, Zhihao and Zhu, Chunyang and Qiu, Jiaxing and Yan, Yuchen and Liu, Jiyuan and Tang, Wenhao and others},
  journal={arXiv preprint arXiv:2607.08448},
  year={2026}
}

@inproceedings{sam3,
  title   = {{SAM} 3: Segment Anything with Concepts},
  author={Carion, Nicolas and Gustafson, Laura and Hu, Yuan-Ting and Debnath, Shoubhik and Hu, Ronghang and Suris Coll-Vinent, Didac and Ryali, Chaitanya and Alwala, Kalyan Vasudev and Khedr, Haitham and Huang, Andrew and others},
  booktitle={International conference on learning representations},
  year={2026}
}

@inproceedings{ames2019cbf,
  title     = {Control Barrier Functions: Theory and Applications},
  author    = {Ames, Aaron D. and Coogan, Samuel and Egerstedt, Magnus and Notomista, Gennaro and Sreenath, Koushil and Tabuada, Paulo},
  booktitle = {European Control Conference (ECC)},
  year      = {2019}
}

@article{brunke2025semantic,
  title   = {Semantically Safe Robot Manipulation: From Semantic Scene Understanding to Motion Safeguards},
  author  = {Brunke, Lukas and Zhang, Yanni and R{\"o}mer, Ralf and Naimer, Jack and Staykov, Nikola and Zhou, Siqi and Schoellig, Angela P},
  journal = {IEEE Robotics and Automation Letters},
  year    = {2025}
}

@article{zhang2026wamvsvla,
  author  = {Zhang, Zhanguang and Li, Zhiyuan and Rahmati, Behnam and Yang, Rui Heng and Ma, Yintao and Rasouli, Amir and Pakdamansavoji, Sajjad and Wu, Yangzheng and Zhang, Lingfeng and Cao, Tongtong and others},
  title   = {Do World Action Models Generalize Better than VLAs? A Robustness Study},
  journal = {arXiv preprint arXiv:2603.22078},
  year    = {2026}
}

@article{meng2025growing,
  title         = {Growing with Your Embodied Agent: A Human-in-the-Loop Lifelong Code Generation Framework for Long-Horizon Manipulation Skills},
  author        = {Meng, Yuan and Sun, Zhenguo and Fest, Max and Li, Xukun and Bing, Zhenshan and Knoll, Alois},
  year          = {2025},
  journal       = {arXiv preprint arXiv:2509.18597},
}

@article{fu2026capx,
  title         = {CaP-X: A Framework for Benchmarking and Improving Coding Agents for Robot Manipulation},
  author        = {Fu, Letian and Yu, Justin and El-Refai, Karim and Kou, Ethan and Xue, Haoru and Huang, Huang and Xiao, Wenli and Wang, Guanzhi and Niu, Dantong and Li, Fei-Fei and others},
  year          = {2026},
  journal={arXiv preprint arXiv:2603.22435},
}

@article{kumar2026actobserve,
  title         = {Act-Observe-Rewrite: Multimodal Coding Agents as In-Context Policy Learners for Robot Manipulation},
  author        = {Kumar, Vaishak},
  year          = {2026},
  journal={arXiv preprint arXiv:2603.04466},
}

@article{xu2026baton,
  title         = {Don't Drop the BATON: Long-Horizon Robot Manipulation via Agentic Subtask Exploration and Transition-aware Memory},
  author        = {Xu, Bingxin and Shang, Yuzhang and Ferrara, Emilio},
  year          = {2026},
  journal       = {arXiv preprint arXiv:2608.16889},
}

@article{haddadin2017collisions,
  title   = {Robot Collisions: A Survey on Detection, Isolation, and Identification},
  author  = {Haddadin, Sami and De Luca, Alessandro and Albu-Sch{\"a}ffer, Alin},
  journal = {IEEE Transactions on Robotics},
  year = {2017}
}

@article{jia2026agentpolicy,
  title   = {Agent as Policy for Robotic Manipulation},
  author  = {Jia, Mengzhao and Lin, Yang and Zhang, Xixin and Zhang, Zhihan and
             Liu, Xiaobai and Jiang, Meng},
  journal = {arXiv preprint arXiv:2609.12541},
  year    = {2026}
}

@article{xiao2026enpire,
  title   = {{ENPIRE}: Agentic Robot Policy Self-Improvement in the Real World},
  author  = {Xiao, Wenli and Xie, Jia and Zhang, Tonghe and Lin, Haotian and Fu, Letian and Xue, Haoru and Lu, Jalen and Yang, Yi and Dai, Cunxi and Wang, Zi and others},
  journal = {arXiv preprint arXiv:2606.19980},
  year    = {2026}
}

@misc{robocurve2026astra,
  title        = {{GPT-6 Astra} on Robotic Manipulation},
  author       = {Menon, Achu and Machcha, Sravanthi and Zou, Sabrina and Chan, Tzu Kit and Chooi, Jay},
  howpublished = {Robocurve evaluation report, \url{https://openai.robocurve.org/gpt-6-astra/}},
  year         = {2026},
  note         = {4 September 2026}
}

@article{chen2026hazardarena,
  title   = {{HazardArena}: Evaluating Semantic Safety in Vision-Language-Action Models},
  author  = {Chen, Zixing and Gao, Yifeng and Wang, Li and Zhao, Yunhan and Liu, Yi and Li, Jiayu and Zheng, Xiang and Wu, Zuxuan and Wang, Cong and Ma, Xingjun and Jiang, Yu-Gang},
  journal = {arXiv preprint arXiv:2604.12447},
  year    = {2026}
}

@article{yang2026saferelbench,
  title   = {{SafeRelBench}: A Spatial-Relation-Aware Benchmark for Process-Level Safety in {VLM}-Driven Embodied Agents},
  author  = {Yang, Huaigang and Li, Ya and Ren, Min and Dai, Bo and Zhang, Zhenliang and He, Zhaofeng},
  journal = {arXiv preprint arXiv:2607.14543},
  year    = {2026}
}

@misc{openai2026astra,
  title        = {{GPT-6} {Astra}: A New Generation of Intelligence},
  author       = {{OpenAI}},
  howpublished = {\url{https://openai.com/index/gpt-6-astra/}},
  year         = {2026},
  note         = {Accessed 2026-09-14}
}

@inproceedings{lu2026isbench,
  title   = {{IS-Bench}: Evaluating Interactive Safety of {VLM}-Driven Embodied Agents in
             Daily Household Tasks},
  author  = {Lu, Xiaoya and Chen, Zeren and Hu, Xuhao and Zhou, Yijin and Zhang, Weichen and
             Liu, Dongrui and Sheng, Lu and Shao, Jing},
  year    = {2026},
  booktitle={Proceedings of the AAAI Conference on Artificial Intelligence},
}

@misc{su2026astra,
  title        = {{GPT 6 Astra} as an Embodied Policy},
  author       = {Su, Jiayi and Zheng, Yixin and Yan, Mi and Yi, Li and Zhang, Zhizheng and Wang, He},
  year         = {2026},
  howpublished = {Technical report and code},
  url          = {https://github.com/anonymous-report-421/eval-of-gpt-6-astra-as-policy}
}

@inproceedings{du2023unipi,
  author        = {Du, Yilun and Yang, Mengjiao and Dai, Bo and Dai, Hanjun and Nachum, Ofir and
                   Tenenbaum, Joshua B. and Schuurmans, Dale and Abbeel, Pieter},
  title         = {Learning Universal Policies via Text-Guided Video Generation},
  booktitle     = {Advances in Neural Information Processing Systems (NeurIPS)},
  year          = {2023},
}

@inproceedings{wu2024gr1,
  author        = {Wu, Hongtao and Jing, Ya and Cheang, Chilam and Chen, Guangzeng and Xu, Jiafeng and
                   Li, Xinghang and Liu, Minghuan and Li, Hang and Kong, Tao},
  title         = {Unleashing Large-Scale Video Generative Pre-training for Visual Robot Manipulation},
  booktitle     = {International Conference on Learning Representations},
  year          = {2024},
}

@article{cheang2024gr2,
  title         = {{GR-2}: A Generative Video-Language-Action Model with Web-Scale Knowledge for
                   Robot Manipulation},
  author        = {Cheang, Chi-Lam and Chen, Guangzeng and Jing, Ya and Kong, Tao and Li, Hang and Li, Yifeng and Liu, Yuxiao and Wu, Hongtao and Xu, Jiafeng and Yang, Yichu and others},
  year          = {2024},
  journal={arXiv preprint arXiv:2410.06158},
}

@article{cen2025worldvla,
  title         = {WorldVLA: Towards Autoregressive Action World Model},
  author        = {Cen, Jun and Yu, Chaohui and Yuan, Hangjie and Jiang, Yuming and Huang, Siteng and Guo, Jiayan and Li, Xin and Song, Yibing and Luo, Hao and Wang, Fan and others},
  year          = {2025},
  journal={arXiv preprint arXiv:2506.21539},
}

@article{assran2025vjepa2,
  title         = {{V-JEPA 2}: Self-Supervised Video Models Enable Understanding, Prediction and
                   Planning},
  author        = {Assran, Mido and Bardes, Adrien and Fan, David and Garrido, Quentin and Howes, Russell and Muckley, Matthew and Rizvi, Ammar and Roberts, Claire and Sinha, Koustuv and Zholus, Artem and others},
  journal={arXiv preprint arXiv:2506.09985},
  year={2025}
}

@article{lu2026aspire,
  title         = {ASPIRE: Agentic Skills Discovery for Robotics},
  author        = {Lu, Runyu and Wu, Yubo and Kou, Ethan and Fu, Letian and Xiao, Wenli and Mandlekar, Ajay and Xu, Yinzhen and Shi, Guanya and Goldberg, Ken and Chen, Ang and others},
  year          = {2026},
  journal={arXiv preprint arXiv:2607.00272},
}

@article{wang2024codeact,
  author        = {Wang, Xingyao and Chen, Yangyi and Yuan, Lifan and Zhang, Yizhe and Li, Yunzhu and
                   Peng, Hao and Ji, Heng},
  title         = {Executable Code Actions Elicit Better {LLM} Agents},
  journal       = {arXiv preprint arXiv:2402.01030},
  year          = {2024},
}

@article{wang2023voyager,
  author        = {Wang, Guanzhi and Xie, Yuqi and Jiang, Yunfan and Mandlekar, Ajay and
                   Xiao, Chaowei and Zhu, Yuke and Fan, Linxi and Anandkumar, Anima},
  title         = {Voyager: An Open-Ended Embodied Agent with Large Language Models},
  journal       = {arXiv preprint arXiv:2305.16291},
  year          = {2023},
}

@article{liu2023lostmiddle,
  title   = {Lost in the Middle: How Language Models Use Long Contexts},
  author  = {Liu, Nelson F. and Lin, Kevin and Hewitt, John and Paranjape, Ashwin and
             Bevilacqua, Michele and Petroni, Fabio and Liang, Percy},
  journal = {Transactions of the Association for Computational Linguistics},
  year    = {2024},
}

@article{laban2025lost,
  title   = {LLMs Get Lost In Multi-Turn Conversation},
  author  = {Laban, Philippe and Hayashi, Hiroaki and Zhou, Yingbo and Neville, Jennifer},
  journal = {arXiv preprint arXiv:2505.06120},
  year    = {2025}
}

@article{li2024instruction,
  title     = {Measuring and Controlling Instruction (In)Stability in Language Model Dialogs},
  author={Li, Kenneth and Liu, Tianle and Bashkansky, Naomi and Bau, David and Vi{\'e}gas, Fernanda and Pfister, Hanspeter and Wattenberg, Martin},
  journal={arXiv preprint arXiv:2402.10962},
  year={2024}
}

@article{anil2024manyshot,
  title={Many-shot jailbreaking},
  author={Anil, Cem and Durmus, Esin and Panickssery, Nina and Sharma, Mrinank and Benton, Joe and Kundu, Sandipan and Batson, Joshua and Tong, Meg and Mu, Jesse and Ford, Daniel and others},
  journal={Advances in Neural Information Processing Systems},
  year={2024}
}

@inproceedings{wu2025lifbench,
  title     = {LIFBench: Evaluating the Instruction Following Performance and Stability of
               Large Language Models in Long-Context Scenarios},
  author    = {Wu, Xiaodong and Wang, Minhao and Liu, Yichen and Shi, Xiaoming and Yan, He and
               Lu, Xiangju and Zhu, Junmin and Zhang, Wei},
    booktitle={Proceedings of the 63rd Annual Meeting of the Association for Computational Linguistics (Volume 1: Long Papers)},
  year      = {2025},
}
\bibliographystyle{iclr2027_conference}


\end{document}